# Deep Belief Networks used on High Resolution Multichannel Electroencephalography Data for Seizure Detection

**JT Turner, Adam Page, Tinoosh Mohsenin and Tim Oates**
CSEE Department, University of Maryland Baltimore County

## Abstract

Ubiquitous bio-sensing for personalized health monitoring is slowly becoming a reality with the increasing availability of small, diverse, robust, high fidelity sensors. This oncoming flood of data begs the question of how we will extract useful information from it. In this paper we explore the use of a variety of representations and machine learning algorithms applied to the task of seizure detection in high resolution, multichannel EEG data. We explore classification accuracy, computational complexity and memory requirements with a view toward understanding which approaches are most suitable for such tasks as the number of people involved and the amount of data they produce grows to be quite large. In particular, we show that layered learning approaches such as Deep Belief Networks excel along these dimensions.

## 1. Introduction

Evidence based, personalized health care depends crucially on large volumes of data about both individuals and populations. It is easy to imagine a near future in which it is common to wear a number of bio-sensors that continuously monitor various aspects of our physiological state, including heart rate, blood pressure, eye movement, brain activity, and many others. Indeed, Qualcomm's recently announced Tricorder XPRIZE is offering $10 million for a team that can produce a small device that monitors health state and successfully identifies the existence of a variety of conditions.

There are two aspects of this enterprise - gathering the data and doing something useful with it. Our starting point is the data, and we ask how it is possible to accurately and efficiently extract information from it for purposes of identifying health states. This leads to the related issues of how to represent large volumes of medical time series so that the information they carry about health state is exposed, and what algorithms are best to extract that information. In this paper we focus on these issues in the context of seizure detection.

In a clinical setting, electroencephalography (EEG) can be used to survey electrical activity in the brain, which can be used to diagnose and monitor abnormal brain functioning. EEGs are often used to diagnose certain neurological conditions such as seizures. Automated seizure detection is still a difficult task, and often produced false positives. In their current state, automated EEGs are not accurate enough for usage in a clinical setting.

Time series are an appropriate model for this problem because of the nature of waveform data collected from an EEG. While the data is often shown as continuous wave forms, which is how humans are able to accurately detect seizures and identify other brain activities from the EEG readings, the data that is received by the machine itself is many discrete electrical readings measured in millivolts (mV). Depending on the design of the actual system itself, the number of readings per second (Hz) varies (the high resolution clinical EEG that is used in the experiment measures at 256 Hz, many commercial EEGs designed for brain-computer interfacing measure at 128 Hz), making time series analysis techniques appropriate for the task.

In this study we consider the problem of detecting whether a patient is having a seizure or not based upon the patients EEG readings for any given second, and how those readings differ from a baseline that is standardized from either the patient's EEG history or other patient's EEG readings.

The remainder of this paper is organized as follows: Section 2 provides background information about the problem of automated seizure detection with special emphasis on time series and deep belief networks, and also discusses related works in the field of time series and deep learning around EEG signals. Section 3 is a detailed overview of the dataset used, a mathematical justification for feature set used, and a description of the Deep Belief Networks used. Section 4 discusses results obtained from the study, and analyzes the results along with complexity and memory requirement discussion, and Section 5 concludes the study.

## 2. Background and Related Work

Time series are prevalent in diverse domains such as finance, medicine, industrial process control, and meteorology. One widely used technique for representing time series is Symbolic Aggregate approXimation (SAX), which converts real-valued data to a sequence of symbols (Lin et al., 2007). More recently, deep learning has shown great promise in tasks such as robotic vision and data mining (Bengio, 2009). With the use of graphics processing units (GPUs) it is possible to train deep artificial neural networks in a layer wise fashion to tackle problems that previously required discretization. In the remainder of this section we introduce terminology, review the deep learning methods used in this work, and discuss related work in the domain of seizure detection using machine learning.

### 2.1. Multichannel Time Series

Let $T$ be a time series representing an EEG that samples at a rate of $H$ Hz over $C$ channels for $S$ seconds. The multichannel time series can be denoted as follows (where $X_{c,s,h}$ is the reading of the $c^{th}$ channel on the $h^{th}$ sample of the $s^{th}$ second, measured in mV):

$$T = \Big(((x_{0,0,0}, \quad x_{0,0,1}, ..., x_{0,0,H-1}),$$
$$(x_{1,0,0}, \quad x_{1,0,1}, ..., x_{1,0,H-1}),$$
$$...$$
$$(x_{C-1,0,0}, \quad x_{C-1,0,1}, ..., x_{C-1,0,H-1})),$$
$$...$$
$$((x_{0,S-1,0}, \quad x_{0,S-1,1}, ..., x_{0,S-1,H-1}),$$
$$(x_{1,S-1,0}, \quad x_{1,S-1,1}, ..., x_{1,S-1,H-1}),$$
$$...$$
$$(x_{C-1,S-1,0}, \quad x_{C-1,S-1,1}, ..., x_{C-1,S-1,H-1}))\Big).$$

This results in a massive amount of high resolution clinical quality EEG data that is so large that it is both unwieldy and may dilute information critical to the task of seizure detection. Featurization of the raw data is critical and is discussed in Section 3.

### 2.2. Classifiers used in this work

Three classifiers are used in this work to compare the detection accuracy and computational and memory requirements. These classifiers are: K-nearest neighbor (KNN) with 3, 5, and 7 neighbors, Support Vector Machines (SVM) with sigmoid, radial basis function, and polynomial kernels, and logistic regression. Figure 1 shows a schematic description of these three classifiers.

### 2.3. Deep Belief Networks

Deep Belief Networks are a new algorithm in the deep learning family, consisting of a set of $N$ Restricted Boltzmann Machines (RBMs) $R = r_0, r_1, ..., r_{N-1}$ that are paired with a classifier $C$ that takes as input the output of the final RBM and outputs class probabilities. The RBMs are stacked in such a way that the output layer of the RBM at layer $\ell - 1$ is the input to the RBM at layer $\ell$. A deep belief network is a system of 2 or more RBMs stacked in this way.

The purpose of the RBM is to learn a probability distribution over a set of inputs, and each RBM consists of the following components: a weight matrix $W$ of size $i \times j$, where $i$ is the number of visible nodes and $j$ is the number of hidden nodes, a visible bias $v$ of length $i$, and a hidden bias $h$ of length $j$. Each RBM is in essence a bipartite graph, with one group of nodes being the visible layer of the RBM, and the other group of nodes being the hidden layer of the RBM.

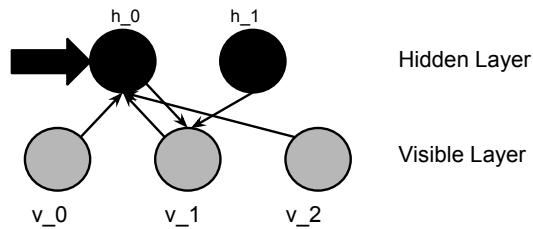

*Figure 2.* The hidden node pointed to by the arrow is a column of the $3 \times 2$ weight matrix $W$, and a scalar of the weight vector $h$.

The concept of a node in an RBM is an abstraction to help us visualize them. A "hidden node" is just the column $W_j$ of $W$, and the scalar $h_j$ of $h$; similarly, a "visible node" is the row $W_i$ of $W$, and the scalar $v_i$ of $v$.

What the RBM does is to learn the probability distri-bution of the inputs. It does this in a stochastic manner such that the machine is less likely to be entrapped on local maxima. Two other important components of RBMs are that they are able to facilitate dimensionality reduction or expansion as is needed for the proper level of abstraction, and that they can be trained in a greedy manner one layer at a time.

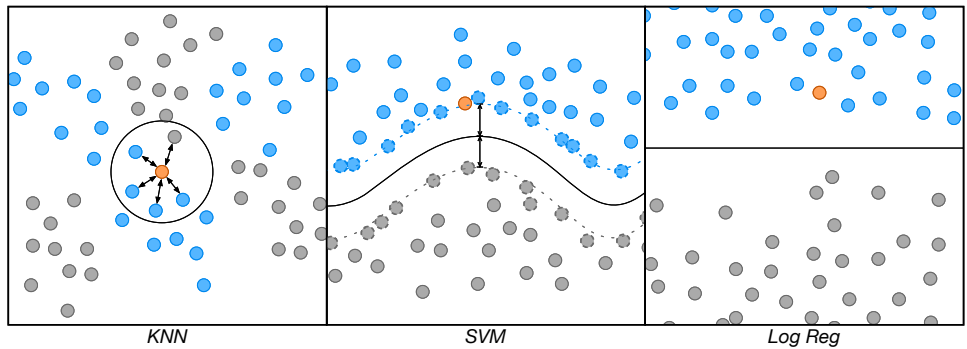

*Figure 1.* The k-Nearest Neighbor (KNN) classifier labels an instance by a majority vote of its $k$ closest neighbors in the instance space according to some distance function. Support vector machines (SVM) learn linear separators in high-dimensional spaces that depend on the kernel used such that the separators are often non-linear in the original feature space. Logistic regression (LR)learns a weight vector that maps instances to class probabilities through a logistic function which, when thresholded, results in a linear separator.

The ability to change the dimensionality of the input is a powerful idea, and the basis for the success deep learning methods have in such areas as robotic vision. DBNs morph the input space into a larger space where abstract objects such as lines, edges, and corners can be formed, and then furthermore constricting that space tighter so that high level objects such as smooth valleys, peaks, and flat lines can be observed by the machine. The artificial neural network generated by the stacking of RBMs is capable of taking raw data (pixels, or in our case electrical signals), and transforming it into data that is useful for the machine to use and process.

Training deep belief networks is best done one layer at a time, in a layerwise manner (Bengio et al., 2007). RBM $r_0$ can be completely trained to gather the structure of the underlying data independently of $r_1, r_2...$, by taking the difference of the *positive gradient* (which is obtained by computing the outer product of the visible input sample $v$, and the sampled hidden layer $h$), and the *negative gradient* (obtained by computing the outer product of a reconstruction $v'$ (by sampling from $h$), and $h'$ (by sampling from $v'$), and adjusting the matrix $W$ as needed from the difference of these gradients.

Once every layer of the deep belief network has been trained using the method described above, the output layer of the final RBM can be used as the input to a classifier. For this study, a simple logistic regression classifier was used, although an SVM or kNN could easily be used in its place.

### 2.4. Related Work

This study builds upon previous studies in the area of seizure detection, deep belief networks, and time series analysis of high resolution medical data.

In a study by Wulsin (2011), deep belief networks were also used for analysis of data obtained from an EEG. The feature set that we chose to use was borrowed from a larger set of features used in this study, however this study attempted to classify anomalous EEG features such as GPED, PLED, or eye blinks as opposed to seizure detection.

A particularly useful study by Shoeb and Guttag used the same dataset of seizure patients that were being monitored by high resolution EEGs after being withdrawn from anti-seizure medications (2009). Although using the same dataset, the Shoeb study extracted a different feature set, and used a support vector machine as the binary classifier, as opposed to a deep belief network. Furthermore, in this study the seizure progression was not interrupted, and statistics were kept on not only the accuracy of seizures detected, but the amount of time that was taken to detect the seizures by the support vector machine.

A final study by Oates et al. (2012) motivated this study and paper. The paper did not study seizure detection, rather traumatic brain injury outcomes. The Oates study investigated time series of high resolution medical data as well, however the data in this study was pulse rate, and $SpO_2$ levels. The study used a Bag of Patterns approach to pre-process data to be used in 1NN clustering to clasify early outcome predictions of patients with traumatic brain injuries.

## 3. Method and Approach

Because using the raw signal input as the input to the deep belief network or classifiers does not allow for the algorithm to properly abstract from the raw data, certain features of the dataset are derived from the raw time series signal. Because a trained human can look at the EEG wave pattern and determine whether or not a seizure is occurring with close to perfect accuracy, many of the features extracted are visible features of the time series such as area under curve, or variation of peaks. The following features were used for detection of anomalous EEG features in the Wulsin study (2011).

### 3.1. Features used

In the following definitions, a **peak** is defined as a reading that marks the change from a positive to negative derivative, and a **valley** is defined as a reading that marks a change from a negative derivative to a positive derivative. $k_i$ will mark the index of the $k^{th}$ peak of the time series, with a value of $x_{k(i)}$. Similarly, $v_i$ will mark the $i^{th}$ valley index, and $x_{v(i)}$ will mark the value of it. Window size is given by $W$. Given a single channel time series $T$, with reading $x_i \forall i \in T$:

- *Area*: Area under the wave for the given time series. Computed as:

$$A = \frac{1}{W} \sum_{i=0}^{W-1} x_i.$$

- *Normalized Decay*: Chance corrected fraction of data that has a positive or negative derivative. $I(x)$ is a boolean indicator function, whose value is 1 when true, 0 when false.

$$D = |\frac{1}{W-1} \sum_{i=0}^{W-2} I(x_{i+1} - x_i < 0) - .5|.$$

- *Line Length*: Summation of distance between all consecutive readings.

$$\ell = \sum_{i=1}^{W} -1|x_i - x_{i-1}|.$$

- *Mean Energy*: Mean energy of time interval.

$$E = \frac{1}{W} \sum_{i=0}^{W-1} |x_i^2|.$$

- *Average Peak Amplitude*: Log base 10 of mean squared amplitude of the $K$ peaks.

$$P_A = \log_{10} \Big( \frac{1}{K} \sum_{i=0}^{K-1} x_{k(i)}^2 \Big).$$

- *Average Valley Amplitude*: Log base 10 of mean squared amplitude of the $V$ valleys.

$$V_A = \log_{10} \Big( \frac{1}{V} \sum_{i=0}^{V-1} x_{v(i)}^2 \Big).$$

- *Normalized Peak Number*: Given $K$ peaks, normalized peak number is the number of peaks normalized by the average difference between data readings.

$$N_P = K \Big( \frac{1}{W-1} \sum_{i=0}^{W-2} |x_{i+1} - x_i| \Big)^{-1}.$$

- *Peak Variation*: Variation between peaks and valleys across time (measured in Hz), and electrical signal (measured in mV). In the case where the number of peaks is not equal to the number of valleys (if an time interval begins or ends during an increase or decrease in data, it is not recorded as a peak or valley), than the feature with the least features is used in comparisons between peaks and valleys. The mean ($\mu(PV)$) and standard deviation ($\sigma(PV)$) of the indicies are given by:

$$\mu(PV) = \frac{1}{K} \sum_{i=0}^{K-1} K_i - V_i.$$

$$\sigma(PV) = \sqrt{\frac{1}{K-1} \sum_{i=0}^{K-1} (K_i - V_i - \mu(PV))^2}.$$

The difference in readings is given by

$$\mu(x_{PV}) = \frac{1}{K} \sum_{i=0}^{K-1} x_{k(i)} - x_{v(i)}.$$

$$\sigma(x_{PV}) = \sqrt{\frac{1}{K-1} \sum_{i=0}^{K-1} (x_{k(i)} - x_{v(i)} - \mu(x_{PV}))^2}.$$

The peak variation is calculated as

$$P_V = \frac{1}{\sigma(PV)\sigma(x_{PV})}.$$

- *Root Mean Square*: The square root of the mean of the data points squared.

$$R_{MS} = \sqrt{\frac{1}{W} \sum_{i=0}^{W-1} x_i^2}.$$

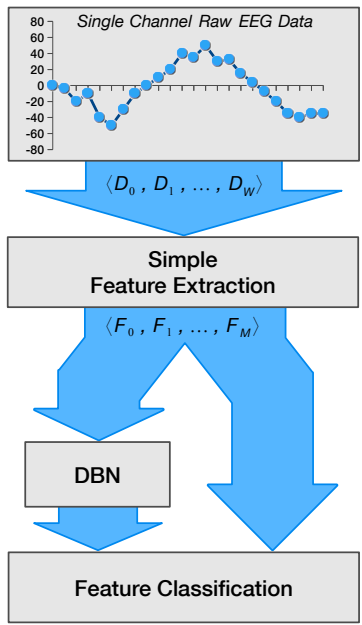

*Figure 3.* Flow diagram of our experiments for seizure detection using two parts.

These nine features were used as the feature space for a single channel of EEG input. In this study, the high resolution EEG data received from 23 channels, so after featurization, the size of the input vector was $23 * 9 = 207$ real numbered values. Before the raw data was converted to features, the data was normalized, with $\mu = 0, \sigma = 1$, and the top 2.5% and bottom 2.5% of values were truncated to 2 and -2 respectively. The normalization was done with respect to each channel individually. Features were then calculated, and then were standardized from $[0, 1]$ for the final input to the deep belief network or a classifier. We used two different approaches to investigate the seizure detection accuracy; Part 1 which uses simple features extraction followed by three different classifiers: SVM, KNN and logistic regression. Part 2 uses simple features extraction followed by DBN and a classifier which is logistic regression. Figure 3 shows the flow diagram for our experiment approach, where $D_i$ represents the digitized raw EEG data within a 256 window size ($W$=256) and $F_i$ represents the corresponding feature for a single channel ($N = 9$ features).

### 3.2. Design & Parameters for DBN

The deep belief network training program was obtained from the Theano library from the LISA lab of University of Montreal (Bergstra et al., 2010), with modifications made to save best models (not only most recent), and improved methods to allow training progress to be monitored. Training was done for a significant amount of time to enhance results, which was possible using GPU calculations which proved to be much faster than using the CPU alone. The calculations were done on a Dell Precision M4700 model with an Intel Core i7-3940XM CPU @ 3.00GHz, 16 GB of memory @ 1866 MHz, and the graphics card used for GPU calculations was an NVIDIA Quadro K2000M with 384 unified pipelines @ 745 MHz, with 2 GB of video memory.

The input data was changed to the features shown in Secion 3. The EEG that collected the readings sampled from 23 channels at 256 hertz, so using the raw data as input to the algorithm would prove difficult, as each second would contain 5,888 distinct EEG readings for the machine to process.

The number of input nodes to the deep belief network were set at 207, with 2 output nodes to classify a second of EEG data as a seizure or non seizure. The number of layers, and number of nodes inside each of the hidden layers of the RBMs was determined through extensive trial and error. The best parameter set was found to be two hidden layers of 500 nodes each. Using CD-1 was found to be sufficient to the task of building structure for classification, and 25 epochs were performed on each layer of the RBM in pretraining, with a learning rate $\alpha = .001$. After the pretraining process of abstraction was completed (without the usage of class labels), the logistic regression layer was trained in the finetuning process. 16 iterations of finetuning were completed, with a learning rate $\alpha = .1$.

Code for classifying instances given a training model was trivial, and emphasizes one of the highlights of deep belief networks. The input to the level $k$ of the network was multiplied by the weight matrix $W_k$, and added to bias matrix $W_b$. The hidden bias and inverted weight matrices used in the pretraining algorithm are not used, because for classification we do not wish to introduce noise into our sample by repropogating the input. When the final layer of the DBN is reached, the argmax of the output layer is taken to assign the class label.

## 4. Results and Analysis

We used two different approaches to investigate the seizure detection accuracy: part 1 which uses simple features extraction followed by three different classifiers: SVM, KNN and logistic regression. Part 2 uses simple features extraction followed by DBN and a classifier, which is logistic regression in this case.

In addition, two different methods of classification tasks were done on the data. In one study the same patient was used for both training, validation, and test-ing sets. This led to a much smaller corpus, but had very good results. The second study involved using all of the other nine patients with data for training and validation sets, and then using one patient at a time for a testing set. This allowed for a much larger corpus for training and testing, but did not produce results as high as the first study. In every study, the majority of the time, the patient was not having a seizure (the fraction of non seizure time varies between 85 - 99 % of the time of the ten patients), so if the only metric used were classification accuracy, a majority class label selection algorithm would achieve accuracy $\geq 85\%$ on every set. For this reason, the metrics of precision, recall, and F-Measure ($F_1$) were used.

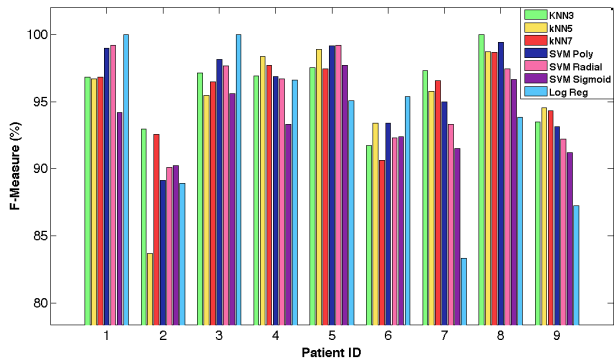

*Figure 4.* Comparison of different classifiers when single patient data is used for training and test

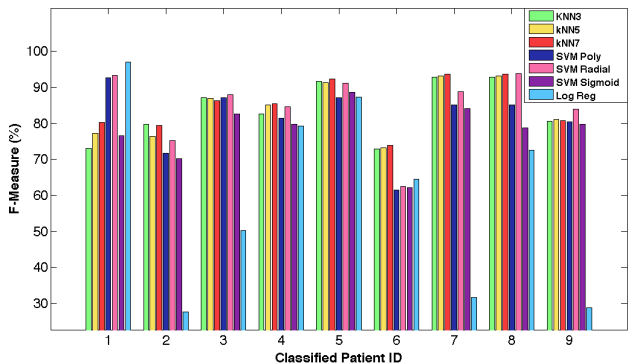

*Figure 5.* Comparison of different classifiers when other patients data are used for training

### 4.1. Part 1: Simple Feature to Classifiers Comparison

#### 4.1.1. $F_1$ AND ACCURACY MEASUREMENTS

In the first study, the training, validation, and testing sets were all drawn from the same patient. The fraction of total seconds to each of the sets are as follows: 71.4% training set, 14.2% validation set, 14.2% testing set. These fractions are derived from the MNIST digit classification method of using a 5:1:1 ratio.

The bar plots in Fig. 4 show the $F_1$ comparison between classifiers when single patient data is used for training and test. In the second study, the training and validation sets were split amongst all of the seizure and non seizure seconds from the nine patients **not** being tested on, using a 4:1 ratio. For the test patient, all of his seizure and non seizure seconds were used in the testing set (since no training or validation was done on the test patient). The bar plots in Fig. 5 show the F1 comparison between classifiers when the patient data is left out for training and is used for testing.

#### 4.1.2. COMPUTATIONAL AND MEMORY COMPLEXITY REQUIREMENTS

Besides the ability for the classifiers to accurately predict seizures, it is also necessary for the classifiers to minimize complexity since they will be running on a low-power, embedded sensor device in ambulatory setting. Since the device can be trained offline, the complexity comes in the form of memory required to store the classifier model and computation required to classify an incoming test vector. Table 1 summarizes the memory and computational complexity for each of the classifiers. The memory and computation required for all the simple features is denoted as SF. Also included in the table is condensed nearest neighbor, CNN. CNN is an optimization applied to KNN that attempts to remove low-content model data while maintaining nearly the same accuracy. The variables used in the table are defined below:

- $W = Window\ Size\ \ (256)$
- $T = \#\ Training\ Windows\ \ (10,000)$
- $C = \#\ Channels\ \ (23)$
- $M = \#\ Features/Channel\ \ (9)$
- $R = Bit\ Resolution\ \ (32)$
- $N = \#\ Neighbors\ \ (5)$
- $L = \#\ DBN Number\ of\ layers\ \ (2)$
- $\alpha_K = Peak\ Ratio\ \ (0.125)$
- $\alpha_{CNN} = CNN\ Reduction\ Ratio\ \ (0.25)$
- $\alpha_{SVM} = SVM\ Support\ Vector\ Ratio\ \ (0.05)$

To better understand how each classifier does relative to one another experimental values were assigned to each variable. These values are shown in parenthesis next to each variable. The last two columns show the relative memory and computation requirements, respectively, for each classifier relative to logistic reqression, which did the best for both requirements. KNN did by far the worst for both cases. This makes sense since KNN requires storing all of the unique training data and labels. For the experimental values, KNN required 10,000x more memory and over 1,000x more computations than logistic regression. For CNN, experimental results showed a reduction of roughly 75% relative to KNN ($\alpha_{CNN} = 0.25$), with only a 5% hit in accuracy. Therefore, it makes sense that CNN requires 2,500x more memory and roughly 275x more computations than LR. SVM did the second best requiring roughly 500x more memory and almost equal amount of computation compared to LR.

| Classifier | Memory Requirement | Computation Requirement | Memory Req. Relative to LR | Computation Req. Relative to LR |
|---|---|---|---|---|
| SF | 0 | $19W + 16\alpha_K W + 10$ | - | - |
| KNN | $TR(CM+1)$ | $3T(CM+N) + (N+1) + SF$ | 10,000x | 1,096.5x |
| CNN | $\alpha_{CNN}TR(CM+1)$ | $3\alpha_{CNN}T(CM+N) + (N+1) + SF$ | 2,500x | 274.8x |
| SVM | $\alpha_{SVM}TR(CM+2)$ | $2CM + \alpha_{SVM}T + 5 + SF$ | 502.5x | 1.086x |
| LR | $R(CM+2)$ | $2CM + 5 + SF$ | 1x | 1x |

*Table 1.* Comparison of memory and computational complexity requirements for simple feature extraction (SF), KNN, CNN, SVM and LR classifiers. The last two columns show the relative memory and computation requirements, respectively, for each classifier relative to logistic reqression, which did the best for both requirements.

### 4.2. Part 2: Simple Feature to DBN and Classifier Comparison

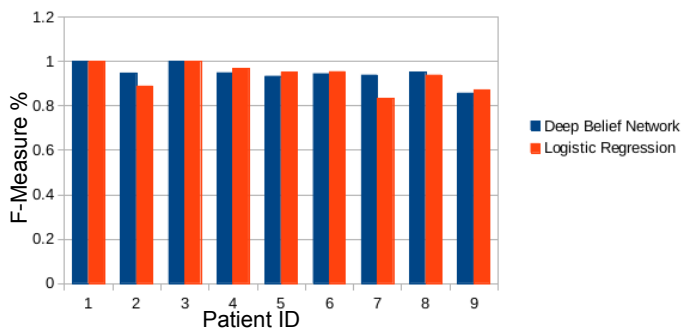

*Figure 6.* Comparison of DBN and logistic regression detection accuracy (F1 score) with single Patient Testing. DBN shows some, but not significant improvement used in single patient training and testing

Since Logistic Regression performs very well both in terms of accuracy and complexity requirements, we used it as the classifier for DBN analysis. Similar to Part 1, we performed the test on single patient training, as well as leaving one patient out for training.

#### 4.2.1. F1 AND ACCURACY MEASUREMENTS

Classification using the same patient as the training and testing corpus is generally an easier task for machines to learn on, so the differences between the deep belief network and the logistic regression are not as great on single patient training as the next study of leave one out training. The deep belief network algorithm was very effective at detection, with two perfect $F_1$ measures, and only one $F_1$ measure below 0.9. These same tests were also run against the same implementation of logistic regression that is used in the output layer of the deep belief network, with f score comparisons shown in Figure 6.

In the second study similar to Part 1, the patient data was left out for training and was used for testing only. $F_1$ measures were lower in this study as was expected, because the test set was similar, but not identical to the sets that the model was trained with, nor validated with. In this second study, 1 patient was above .9, 4 patients were between .8 and .9, and only 3 patients were below .8. Compared against the same implementation of logistic regression that takes the output layer of the network as input run by itself, the results are shown in Figure 7. In this harder machine learning problem of leave one out patient training, the deep belief network shows much improved performance over the logistic regression algorithm. Although the improvement is not better in all nine patients, in many of them there is a very significant improvement in classification $F_1$ measure from the deep belief network.

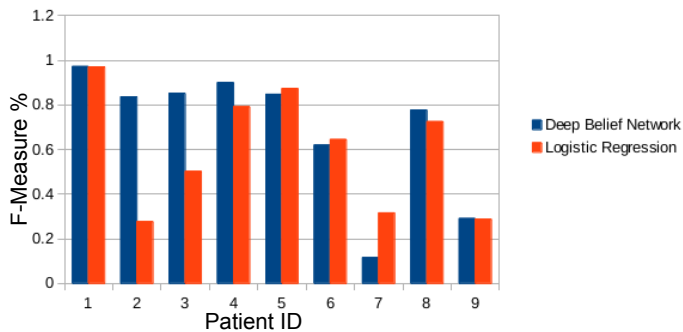

*Figure 7.* Leave one out patient training. The Deep Belief Network shows significant improvements over logistic regression in many of the cases.

#### 4.2.2. Computational and memory complexity requirements

As discussed previously in section 4.1.2, complexity of the system must also be examined. Adding a DBN stage into the system will increase both the memory and computation. In terms of storage, a DBN stage will add approximately $LR(CM)^2$ more bits than just logistic regression, where L is the number of layers. This is assuming that the average number of nodes in a layer is equal to the number of input features. For our experiments, this required 413x more memory than LR. In terms of complexity, the DBN stage will add approximately $LCM(2CM+1)$. Again, from our experiments this required 30x more computations than LR alone.

## 5. Conclusion

In this paper, the use of a variety of representations and machine learning algorithms was applied to seizure detection in high resolution and multi-channel EEG data. Classification accuracy, computational complexity and memory requirements are explored with the view of processing large patient data requirements. Among classifiers logistic regression performs best in terms of complexity and accuracy for the majority of tests. Also, seizure detection in the studies where the same patient was used in the training, validation, and testing sets was very successful on all patients. Although these are good numbers, it may not always be feasible to have hours of trained data about a patient to use as a model. The more realistic clinical study is the study, where the patients tests were done without any previous knowledge of the patient being tested on. Dealing in the domain of using models of other patients to represent a different patient being tested upon (as was the case in the leave one out training and in real situations), deep belief networks often outperformed the logistic regression algorithm using the same feature set.

## References

Bengio, Yoshua. Learning deep architectures for ai. *Found. Trends Mach. Learn.*, 2(1):1–127, January 2009. ISSN 1935-8237. doi: 10.1561/2200000006. URL http://dx.doi.org/10.1561/2200000006.

Bengio, Yoshua, Lamblin, Pascal, Popovici, Dan, and Larochelle, Hugo. Greedy layer-wise training of deep networks. In Schölkopf, Bernhard, Platt, John, and Hoffman, Thomas (eds.), *Advances in Neural Information Processing Systems 19 (NIPS'06)*, pp. 153–160. MIT Press, 2007. URL http://www.iro.umontreal.ca/~lisa/pointeurs/BengioNips2006All.pdf.

Bergstra, James, Breuleux, Olivier, Bastien, Frédéric, Lamblin, Pascal, Pascanu, Razvan, Desjardins, Guillaume, Turian, Joseph, Warde-Farley, David, and Bengio, Yoshua. Theano: a CPU and GPU math expression compiler. In *Proceedings of the Python for Scientific Computing Conference (SciPy)*, June 2010. Oral Presentation.

Lin, Jessica, Keogh, Eamonn, Wei, Li, and Lonardi, Stefano. Experiencing sax: a novel symbolic representation of time series. *Data Mining and Knowledge Discovery*, 15(2):107–144, 2007. ISSN 1384-5810. doi: 10.1007/s10618-007-0064-z. URL http://dx.doi.org/10.1007/s10618-007-0064-z.

Oates, T., Mackenzie, C.F., Stansbury, L.G., Aarabi, B., Stein, D.M., and Hu, P.F. Predicting patient outcomes from a few hours of high resolution vital signs data. In *Machine Learning and Applications (ICMLA), 2012 11th International Conference on*, volume 2, pp. 192–197, 2012. doi: 10.1109/ICMLA.2012.219.

Shoeb, Ali. *Application of Machine Learning to Epileptic Seizure Onset Detection and Treatment*. PhD thesis, Massachusetts Institute of Technology, Cambridge, September 2009.

Wulsin, D F, Gupta, J R, Mani, R, Blanco, J A, and Litt, B. Modeling electroencephalography waveforms with semi-supervised deep belief nets: fast classification and anomaly measurement. *Journal of Neural Engineering*, 8(3):036015, 2011. URL http://stacks.iop.org/1741-2552/8/i=3/a=036015.